\documentclass[letterpaper, 10 pt, conference]{ieeeconf}  % Comment this line out if you need a4paper

\IEEEoverridecommandlockouts                              % This command is only needed if
\usepackage{amsmath} % assumes amsmath package installed
\usepackage{amssymb}  % assumes amsmath package installed

\usepackage[ruled,vlined]{algorithm2e}
\usepackage{booktabs}
\usepackage{graphicx}
\usepackage{hyperref}
\usepackage[dvipsnames]{xcolor}

\title{\LARGE \bf
From Handheld to Unconstrained Object Detection: a Weakly-supervised On-line Learning Approach
}

\author{Elisa Maiettini$^{*1}$, Andrea Maracani$^{*1,2,3}$ Raffaello Camoriano$^{2}$, Giulia Pasquale$^{1}$, Vadim Tikhanoff$^{4}$,\\ \quad\, Lorenzo Rosasco$^{2,3,5}$ and Lorenzo Natale$^{1}$% <-this % stops a space
\thanks{$^{*}$Equal contribution}
\thanks{$^{1}$Humanoid Sensing and Perception, Istituto Italiano di Tecnologia, Genoa, Italy}
\thanks{$^{2}$IIT@MIT - Laboratory for Computational and Statistical Learning, IIT, Genoa, Italy, and MIT, Cambridge, MA, USA}
\thanks{$^{3}$MaLGa \& DIBRIS, Universit\'a degli Studi di Genova, Genoa, Italy}
\thanks{$^{4}$iCub Tech, Istituto Italiano di Tecnologia, Genoa, Italy}
\thanks{$^{5}$Center for Brains, Minds and Machines, MIT, Cambridge, MA, USA}
}

\begin{document}

\maketitle
\thispagestyle{empty}
\pagestyle{empty}

\newcommand{\elisa}[1]{{\color{Green}#1}}
\newcommand{\lorenzo}[1] {{\color{red}{#1}}}
\newcommand{\raf}[1] {{\color{blue}{#1}}}

%%%%%%%%%%%%%%%%%%%%%%%%%%%%%%%%%%%%%%%%%%%%%%%%%%%%%%%%%%%%%%%%%%%%%%%%%%%%%%%%
\begin{abstract}
Deep Learning (DL) based methods for object detection achieve remarkable performance at the cost of computationally expensive training and extensive data labeling. Robots embodiment can be exploited to mitigate this burden by acquiring automatically annotated training data via a natural interaction with a human showing the object of interest, handheld. However, learning solely from this data may introduce biases (the so-called domain shift), and prevents adaptation to novel tasks.
% These methods have important limitations for Robotics: learning solely on off-line data may introduce biases (the so-called domain shift), and prevents rapid adaptation to novel tasks.
While Weakly-supervised Learning (WSL) offers a well-established set of techniques to cope with these problems in general-purpose Computer Vision, its adoption in challenging robotic domains is still at a preliminary stage. 
In this work, we target the scenario of a robot trained in a teacher-learner setting to detect handheld objects. The aim is to improve detection performance in different settings by letting the robot explore the environment with a limited human labeling budget.
% In this work, we target the scenario of a robot exploring the environment, aiming to improve detection performance with a limited human labeling budget.
%In this work, we investigate how WSL can tackle  weakly-supervised learning can cope with these problems. 
We compare several techniques for WSL in detection pipelines to reduce model re-training costs without compromising accuracy, proposing solutions which target the considered robotic scenario. 
% We show that diversity sampling for constructing Active Learning queries and strong positives selection for Semi-supervised Learning both enable significant annotation savings and improve adaptation to novel domains.
We show that the robot can improve adaptation to novel domains, either by interacting with a human teacher (Active Learning) or with an autonomous supervision (Semi-supervised Learning).
We integrate our strategies into an on-line detection method, achieving efficient model update capabilities with few labels.
%In particular, we show that diversity sampling for constructing active learning queries and strong positives selection for self-supervised learning enable significant annotation savings and improve domain shift adaptation 
%By integrating our strategies into a hybrid DCNN/FALKON on-line detection pipeline \cite{maiettini_weakly_2019}, our method is able to be trained and updated efficiently with few labels, overcoming limitations of previous work.
We experimentally benchmark our method on challenging robotic object detection tasks under domain shift.
Code will be released for reproducibility at camera-ready stage.

\end{abstract}

%%%%%%%%%%%%%%%%%%%%%%%%%%%%%%%%%%%%%%%%%%%%%%%%%%%%%%%%%%%%%%%%%%%%%%%%%%%%%%%%
\section{INTRODUCTION}\label{sec:intro}

In the state-of-the-art, object detection is typically addressed with DL-based approaches~\cite{He2017,yolov4} that achieve remarkable performance. Despite their high accuracy, they are constrained by requiring long training times and large annotated datasets, limiting their adoption in such applied settings where quick adaptation to novel tasks is required. In Robotics, the embodiment of a robotic agent can be exploited to interact with the environment, including humans, to mitigate this burden and actively acquire training data. Regarding the interaction with humans, past work shows that a teacher-learner scenario can be exploited to automatically collect labeled images for object recognition~\cite{pasquale2019} and detection~\cite{maiettini2017}. Specifically, in those works the human teacher shows an object, while holding it in their hand, to the robot and 3D information is used to automatically collect the location information. However, while effective and allowing for a natural interaction, this approach supports limited generalization to novel, unseen, scenarios~\cite{maiettini2017,maiettini2019b}. A further possibility is to exploit robots ability to navigate and autonomously explore the environment, acquiring training images during operation.
Such images come in streams and can carry useful information, eventually containing the objects of interest, but they are not labeled.
\textit{Weakly-supervised Learning} (WSL)~\cite{zhou_brief_2018}, is a well-established general purpose Computer Vision framework which targets learning from partially-annotated datasets. However, despite initial work in robotic vision~\cite{Jamieson2020,maiettini2019b} 
%
% (see e.g.~\cite{fanello2013} for object recognition, [ADD MORE?] and specifically~\cite{maiettini2019b} for object detection), 
%
the robotic literature misses a thorough comparison that investigates advantages and limitations of existing techniques, especially in the context considered in this paper.
%
%they have not still been bench-marked on a typical robotic setting, comprising severe limitations in human labeling budget and highly time correlated frames.
For instance, in~\cite{maiettini2019b}, the unlabeled images are processed with a pre-trained model to either select the hard ones and ask a human expert to help and annotate them (\textit{Active Learning} (AL) framework~\cite{settles_active_2012}) or add the predictions of the easy ones to the training set (\textit{Semi-supervised Learning} (SSL)~\cite{hernandez-gonzalez_weak_2016,zhou_brief_2018}). These frameworks allow for a natural interaction with the environment and the human teacher to improve the visual system and work presented in~\cite{maiettini2019b} effectively reduces the amount of manual annotation, but it has some limitations.
%Promising results have been achieved by integrating the aforementioned on-line object detector with \textit{weakly-supervised learning} (WSL)~\cite{wang_towards_2018}, which  targets learning from partially-annotated datasets. 
%The resulting pipeline exploits the labeled part of the dataset to train an initial \emph{seed} model, which is then used to process the unlabeled images. 
%The obtained predictions are used to form ``hard" and ``easy" image sets. 
%The former is used to construct queries for human annotation (\textit{Active Learning} (AL) framework~\cite{settles_active_2012}), while the second is added to the training set for \textit{Self-supervised Learning} (SSL)~\cite{hernandez-gonzalez_weak_2016,zhou_brief_2018}. 
%While showing promising results, this method presents some limitations. 
Firstly, the unsupervised data 
processing is \textit{pool-based}~\cite{settles_active_2012}, 
that is, all unlabeled images are evaluated before query selection. 
This is not suitable for a robotic system that is exploring the environment and needs to decide interactively whether to request annotations or not. To this aim, \textit{stream-based} techniques~\cite{settles_active_2012} are preferable, because they allow to process images frame by frame and to make individual query decisions on-line. This strategy, however, might yield to lower accuracy since queries are constructed using limited %single-frame 
information on the unlabeled set~\cite{settles_active_2012}. 
Moreover, the pre-trained detection method in~\cite{maiettini2019b} iterates multiple times over the unlabeled data, which, while allowing to refine the data selection, slows down learning. Finally, while succeeding in reducing the human effort required for refinement,~\cite{maiettini2019b} still needs a relatively high number of manual annotations, which prevents its adoption in on-line applications.

In this paper, we study how WSL techniques can be used to exploit the robot interaction with the environment and the human teacher to update and improve performance of object detection models previously trained with data of handheld objects.
We focus on the stream-based scenario with the aim of increasing the human labeling efficiency of weakly-supervised on-line object detection. 
Moreover, we consider the case in which only one pass over the unlabeled data is allowed. 
The main contributions of this work are as follows. We present and empirically evaluate several AL techniques for detection, typically used in general purpose computer vision. We compare \textit{pool-based} and \textit{stream-based} AL in challenging robotic scenarios and propose a solution to overcome limitations of the latter. We also consider the case where no human labeling is allowed for adaptation. Specifically, we investigate the domain shift effects occurring when using a model trained on data of handheld objects in different settings and how wrongly self-annotated data can degrade accuracy in those cases.
% We investigate domain shift effects on AL and SSL and how wrongly self-annotated data can degrade accuracy.
Finally, we propose an SSL sampling method to overcome this problem and we empirically demonstrate that, in case no labeling is allowed, it can effectively improve model performance.
% \begin{itemize}
% 	\item We present and empirically evaluate several AL techniques for detection, typically used in general purpose computer vision. 
% We compare \textit{pool-based} and \textit{stream-based} AL in a challenging robotic scenario and propose a solution to overcome limitations of the latter.
% 	\item We investigate domain shift effects on AL and SSL and how wrongly self-annotated data can degrade accuracy.
% 	\item Finally, we propose an SSL sampling method to overcome this problem and we empirically demonstrate how SSL can effectively reduce labeling efforts.
% \end{itemize}
This paper is organized as follows:
we introduce WSL in Sec.~\ref{sec:bg} and we cover related work in Sec.~\ref{sec:rel}.
In Sec.~\ref{sec:our-approach}, we present our efficient detection methods, which are analyzed and validated in Sec.~\ref{sec:experiments}.
Sec.~\ref{sec:conclusions} concludes the paper.

\section{BACKGROUND}
\label{sec:bg}
% \subsection{Object Detection}
% 
% Given an image 
% $x\in\mathcal{I}$,
% the goal of object detection is to both localize and recognize, in $x$, all the objects belonging to a specified set of $T$ classes, $C$. 
% We define $f_{det}^*: \mathcal{I} \rightarrow \mathcal{D}$ as the ground truth associating $x$ with the $N_x^* \in \mathbb{N}_1$ pairs of class labels $y^*\in{C}$ and locations $z^* \in \mathcal{Z}$ of all the target objects it contains.
% The aim of object detection is to 
% construct a function (detector) $\hat{f}_{det}: \mathcal{I} \rightarrow \mathcal{D}$ approximating $f_{det}^*$ as closely as possible according to a specified quality metric.
% In this work, we consider confidence-based detectors $\bar{f}_{det}: \mathcal{I} \rightarrow \mathcal{D}\times \mathbb{R}$, which associate a confidence score $s_i\in \mathbb{R}$ to each predicted output pair:
% \begin{equation}
% \bar{f}_{det}(x) = \{ ((\hat y_i,\hat z_i),s_i)\}_{i = 1}^{N_x}.
% \end{equation}
% Thresholding $s_i$ with a parameter $\tau\in\mathbb{R}$ allows to 
% select 
% predictions according to the desired trade-off between precision and recall, where $\tau$ is 
% the minimum required confidence score for the detector to predict $(\hat y_i,\hat z_i)$.
% 
% 
% \subsection{From Fully- to Weakly-supervised Learning}
% 
The supervised learning approach to object detection is centered on learning the detector function from an annotated (supervised) dataset $S_n = \{(x_i,Y_i)\}_{i=1}^n$ of images ($x_i$) and corresponding bounding boxes and labels annotations ($Y_i$).
The methods described in Sec.~\ref{sec:rel-generic-approaches} fall in this category. 
They contributed to a clear progress in detection accuracy and prediction speed. However, they need expensively-annotated large-scale datasets to be optimized.
This property does not meet the robotic requirement for a detector to adapt to a variety of tasks, potentially unknown a-priori, in a short time span.
However, while large annotated datasets might not be available, plenty of unsupervised images are usually accessible to robots.
In this context, a training set $S_n = L \cup U$ is typically composed of a labeled subset $L = \{(x_i, Y_i)\}_{i = 1}^{n_L}$
and an unlabeled subset $U = \{x_i\}_{i = 1}^{n_U}$.
WSL allows the agent to select
unsupervised images from $U$ and acquire their labels semi-autonomously for updating the detector, minimizing human effort and improving accuracy.
WSL includes several subclasses of methods, depending on the label-acquisition mechanism \cite{hernandez-gonzalez_weak_2016,zhou_brief_2018}.
The most relevant for this work are Active and Semi-supervised Learning.

\noindent{\bf{Active Learning.}} AL~\cite{settles_active_2012} interactively queries unsupervised examples for expert labeling to minimize human annotation and maximize accuracy.
Unlabeled examples are chosen from $U$ according to a \textit{scoring function} and a \textit{sampling strategy}.
Their labels are then queried to an expert, and newly-annotated examples are added to $L$ for training.
If all images in $U$ are accessible at selection time, sampling is referred to as \textit{pool-based}.
Otherwise, if only one candidate from $U$ is accessible, sampling becomes a binary decision on keeping or dropping it and is called \textit{stream-based}.
The AL selection criterion we focus on is \textit{uncertainty sampling}, which picks the examples the model is \textit{least confident} about.

\noindent{\bf{Semi-supervised Learning.}} In SSL~\cite{zhou_brief_2018}, unlabeled images are annotated by the detector itself with no human intervention, propagating predicted labels to high-confidence regions of the input space by exploiting the geometry of the input data distribution.
This technique is effective if the detector is not overconfident of its predictions and if the confidence threshold for propagating predicted labels is strict enough.

\section{RELATED WORK}
\label{sec:rel}
%In this section, we introduce the main state-of-the-art approaches for general purpose object detection (see Sec.~\ref{sec:rel-generic-approaches}), specifically focusing on the efficiency requirement typical of robotic applications (see Sec.~\ref{sec:rw-efficient-det}).
 
\subsection{Object Detection}\label{sec:rel-generic-approaches}
% Classical
Early approaches to object detection were based on feature dictionaries \cite{papageorgiou_general_1998} or specific kinds of image descriptors \cite{viola_rapid_2001}.
%The extracted 
Feature vectors were separately classified by supervised learning methods.
% such as support vector machines.
Despite yielding limited accuracy,
% for current standards
these approaches had the advantage of being parsimonious in terms of computations and dataset size.
More recently, object detection experienced significant progress thanks to the introduction of DL-based methods.
This determined clear improvements in terms of predictive performance, mainly due to the powerful representation capabilities of deep networks.
% R-CNN
Such approaches include two-stage detectors based on Region Proposal Networks (RPNs)~\cite{ren_faster_2015} (like e.g. Faster R-CNN~\cite{ren_faster_2015} and Mask R-CNN~\cite{He2017}) and related extensions \cite{dai_r-fcn_2016,dai_deformable_2017,pang_libra_2019}.
These methods employ a deep network to perform (i)~region candidates predictions, (ii)~per-region feature extraction and (iii)~region classification and refinement.
% YOLO
%Deep detection methods unifying RPNs and deep classifiers into single architectures trained end-to-end demonstrate further improvements in terms of accuracy and prediction time, allowing for accurate high-frame rate detections \cite{ren_faster_2015,redmon_you_2016}.
% DENSE DETECTORS
Alternative end-to-end approaches include one-stage detectors, which replace the RPN with a fixed, dense grid of candidate bounding boxes.
%, predicting class probabilities and box adjustments.
% SSD
One such example is SSD~\cite{liu_ssd_2016,zhai2020}, achieving accuracies competitive with the RPN-based Faster R-CNN and high frame rate.
% Focal Loss
Another one-stage method, RetinaNet~\cite{lin_focal_2017}, rebalances foreground and background examples through the so-called Focal Loss.
% RefineDet
%RefineDet~\cite{zhang_single-shot_2018} takles the problem of the overwhelming number of negative background examples by introducing a grid pruning module into the one-stage detection pipeline.

\subsection{Efficient Object Detection for Robotics}\label{sec:rw-efficient-det}
% Limitations, especially for robotics
Despite their high accuracy, the approaches described above typically require (i) long training time and (ii) large-scale annotated datasets for adaptation to novel tasks. These aspects limit their adoption in Robotics.

\noindent{\bf{Computational efficiency. }}
%It is desirable for robotic agents to be adaptable to dynamically changing conditions, allowing for seamless human-robot interaction.
%It is well-known that 
A well-known issue of DL-based pipelines is that they suffer from 
%the so-called 
\textit{catastrophic forgetting} when optimized on new data~\cite{goodfellow_empirical_2013}.
This limitation implies retraining these models on the full dataset, causing long adaptation time.
%This technical misalignment calls for the development of more efficient approaches specifically targeted to robotic object detection.
% two-stage pipelines for fast retraining
To address this issue, a recent work 
for robotic object detection 
leverages fast classifiers to enable on-line adaptation~\cite{maiettini2018,maiettini2019a}.
Specifically, in~\cite{maiettini2019a}, an efficient multi-stage pipeline is proposed by combining DL-based RPNs and feature extractors (namely, based on Faster R-CNN or Mask R-CNN) with large-scale Kernel classifiers \cite{rudi_falkon_2017,meanti_kernel_2020,rudi_less_2015}.
According to this approach,
%-- which we follow throughout this work -- 
the feature extractor is pre-trained 
%once and 
off-line on a large representative dataset, yielding a powerful and transferable learned representation, which is kept fixed during on-line operation.
%Later on, the feature extraction and region proposal is kept fixed during on-line operation.
The actual regions classification is performed by the integration of an efficient hard-negatives bootstrapping approach (the Minibootstrap~\cite{maiettini2019a}) with a set of Kernel-based FALKON classifiers \cite{rudi_falkon_2017,meanti_kernel_2020}.
%Since FALKON can be trained quickly,
%the hybrid detection pipeline is suitable for  online operation.
%In particular, within the designed learning protocol~\cite{maiettini_-line_2019}, FALKON can be retrained as soon as data sampled from a new task becomes available, adapting the classifier to domain shift.\\

\noindent{\bf{Labeling efficiency. }}
Labeling efficiency is another key requirement for robotic object detection. 
%Indeed, data labeling is expensive in general, but it can become even unfeasible for training detectors for general-purpose robotic agents.
%The objective is to maximize accuracy while minimizing the number of images for which class and bounding-box labels are queried to a human expert. 
The broad class of WSL methods \cite{hernandez-gonzalez_weak_2016,zhou_brief_2018} provides a rich set of tools towards this goal in general purpose Computer Vision, in particular AL and SSL -- introduced in Sec.~\ref{sec:bg}.
% AL
%Active Learning (AL) \cite{settles_active_2012} selects informative unsupervised examples and queries an oracle (e.g., a human expert) for the corresponding labels.
%The AL selection process 
%(please, refer to Sec.~\ref{sec:bg} for the definition of the framework) 
%is aimed at maximizing accuracy with the minimum budget of queries.
After successful applications to deep object classification~\cite{kirsch_batchbald_2019,zhdanov_diverse_2019,ash_deep_2020-1}, AL has been recently applied also to object detection~\cite{aghdam_active_2019,haussmann_scalable_2020,desai_adaptive_2019}.
% has recently been used in conjunction with DCNN architectures.
For instance, recently, detection-specific image scoring functions (like e.g., \textit{localization tightness} and \textit{stability}~\cite{kao_localization-aware_2019}) have been proposed.
Instead, when no further annotation is allowed to exploit the unsupervised samples, SSL techniques can be used. Similarly to AL, also SSL has been recently applied to object detection.
For instance, in~\cite{li_improving_2020}, SSL is employed for dataset augmentation and training object detectors.
Moreover, the authors point out that vanilla SSL can degrade accuracy in presence of domain shift.
We also observed the same issue in our robotic setting and we propose a simple yet effective solution in Sec~\ref{sec:methods:ss}.
%Importantly, a failure case of
%a SSL object detection method is proposed, employing data augmentation and using a separate network for rectifying augmented data annotations.
%Importantly, the work points out that
%vanilla SSL in presence of domain shift is pointed out.
% Weakly-supervised learning
% SSM & co.
Recent approaches integrate both AL and SSL techniques into the same detection pipeline, such as Self-supervised Sample Mining~\cite{wang_towards_2018,wang_cost-effective_2019} (SSM). 
SSM sorts unsupervised images into separate candidate sets for further AL and SSL processing, according to the predictive confidence scores of the underlying DL-based detection model~\cite{dai_r-fcn_2016}.
Another related field in Computer Vision is \textit{Unsupervised Domain adaptation} for object detection~\cite{oza2021}. Specifically, the \textit{pseudo-labeling} approach~\cite{li2021,chowdhury2019} proposes to adapt a detection model to novel and unknown domains (i.e., datasets) by using confident model predictions as pseudo-ground truth.
%A natural question is whether AL and SSL can both be effectively integrated in the same detection pipeline for gaining further labeling and computational efficency.
%
%\textit{Self-supervised Sample Mining} (SSM)
% \cite{wang_towards_2018,wang_cost-effective_2019} represents the first step in this direction.
%SSM sorts unsupervised images into separate candidate sets for further AL and SSL processing according to the predictive confidence scores of the underlying Region-FCN~\cite{dai_r-fcn_2016} model.
%% 

The aforementioned approaches have been proposed and benchmarked on general purpose Computer Vision datasets. However evaluation of WSL techniques on robotic scenarios is still at an initial stage (e.g., see~\cite{Jamieson2020,maiettini2019b}).
For instance, in~\cite{maiettini2019b}, SSM is extended to enable on-line adaptive object detection for Robotics, by integrating the WSL sample selection strategy with the on-line object detection method~\cite{maiettini2019a}. However,~\cite{maiettini2019b} still requires a relatively large number of manual annotations, does not investigate the effect of severe domain shift in self-supervision and focuses on a \textit{pool-based} processing. While showing encouraging results, all these limitations prevent its adoption in on-line applications. In this work,
%Region-FCN is replaced by the aforementioned on-line hybrid pipeline proposed in \cite{maiettini_speeding-up_2018,maiettini_-line_2019}, while the SSM weakly-supervised sample selection strategy is kept unchanged.
%
%While in~\cite{maiettini_weakly_2019}, the WSL strategy is kept unchanged with respect to SSM, in this work,
%
we present an empirical analysis of different general purpose computer vision AL and SSL techniques in a challenging robotic scenario, targeting a low annotation budget regime. We focus on how WSL techniques can be used to exploit the robot interaction with the environment and the human teacher to update and improve performance of object detection models previously trained with data of handheld objects. Moreover, we propose solutions to overcome the aforementioned limitations, improving the AL performance and addressing the SSL failure cases under domain shift, increasing overall labeling efficiency.

%Cross-Image Validation (CIV) proposed in SSM~\cite{wang_towards_2018}.
%CIV stitches detected image patches on random images sampled from a separate validation set.
%Then, it executed the detection again on the stitched images and evaluates the statistics of the newly-obtained confidence scores. If the predictions pass the validation, they are considered high-confidence and can be used for self-supervision.
%
%\subsection{Contribution overview}
%In this paper, we significantly improve the accuracy and efficiency of \cite{maiettini_weakly_2019} by modifying the weakly-supervised strategy as follows:
%\begin{itemize}
%\item Introducing image-level diversity sampling in the AL selection strategy;
%\item Proposing a novel efficient diversity sampling technique for robotic tasks based on time coherence;
%\item Overcoming the severe accuracy degradation of SSL under domain shift (as also observed in \cite{li_improving_2020}) through a stricter positives selection strategy;
%\item Enforcing diversity also among high-confidence examples in the SSL selection procedure (???).
%\end{itemize}

% methods allowing for online operation

% - Hybrid rfcn-falkon architecture (IROS+AURO paper)

% - Hybrid architecture + weakly supervised (Humanoids paper)

\section{METHODS}
\label{sec:our-approach}
%%%%%%%%%%%%%%%%%%%%%%%%%%%%%%%%%%%%%%%%%%%%%%%%%%%%%%%%%%%%%%%%%%%%%%%%%%%%%%%%

\begin{figure}[t]
	\centering
	\vspace{0.3cm}
	\includegraphics[width=0.37\textwidth]{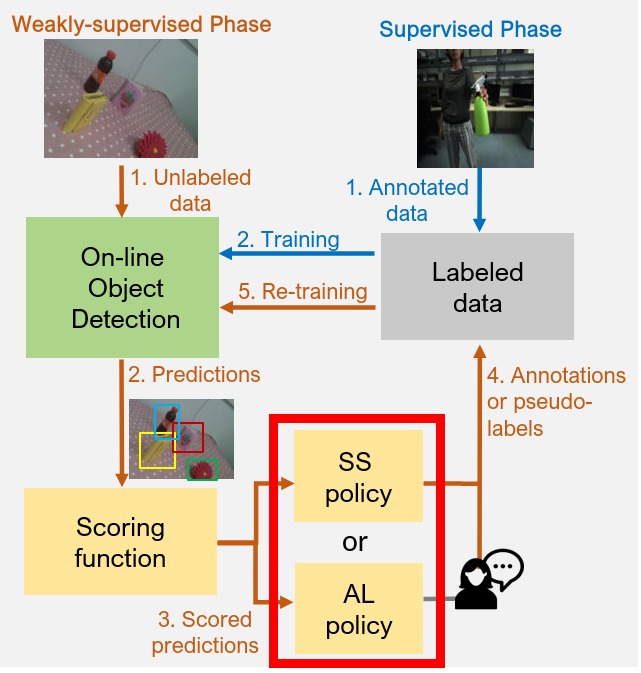}
	\caption{Overview of the proposed pipeline. Refer to Sec.~\ref{sec:methods:overview} for details.}
	\label{fig:pipeline}
\end{figure}
In this work, a robot is asked to detect a set of object instances in an unconstrained environment (referred to as \textsc{TARGET}).
% The convolutional weights of the detection system are initialized with an off-line pre-train on a separate set of objects and 
A first detection model is trained during a brief interaction with a human, in a teacher-learner scenario, like e.g. in~\cite{maiettini2017} where objects are handheld (the \textsc{TARGET-LABELED}). 
Then, the robot autonomously explores the environment, acquiring a stream of images in a new setting, where automatic annotation is not possible. 
Therefore, these images are not labeled (\textsc{TARGET-UNLABELED}) and are used to adapt the detector on-line exploiting the robot interaction with the environment and the human teacher. In the next sections, we present the proposed pipeline (Sec.~\ref{sec:methods:overview}) and the learning protocol (Sec.~\ref{sec:methods:learning}). 
Then, we present all the considered AL and SSL techniques and the proposed approaches (Sec.~\ref{sec:methods:al} and~\ref{sec:methods:ss}, respectively).

\subsection{Pipeline Description}
\label{sec:methods:overview}
The proposed WSL pipeline (see Fig.~\ref{fig:pipeline}) is composed of four main modules: (i)~the \textit{On-line Object Detection}, (ii)~the \textit{Scoring function}, (iii)~the \textit{AL Selection policy}, and  (iv)~the \textit{SS Selection policy}.\\

\noindent{\bf{On-line Object Detection (OOD)}}. 
For this module, we follow the method proposed in~\cite{maiettini2019a}, but considering the implementation presented in~\cite{ceola2020a} and~\cite{ceola2020b}. 
This is an on-line learning approach consisting of two stages: (i)~region proposals and feature extraction, and 
(ii)~region classification and bounding-box refinement.
The first stage relies on layers from Mask R-CNN~\cite{He2017} (specifically, the convolutional layers, the RPN~\cite{ren2015_faster} and the RoI Align layer~\cite{He2017}).
In particular, this part is used to extract a set of Regions of Interest (RoIs) from an image and encode them into a set of features. 
The second stage is composed of a set of FALKON~\cite{rudi_falkon_2017}
binary classifiers (one for each class of the TARGET) and Regularized Least Squares
(RLS)~\cite{hastie_elements_2009}, respectively for the classification and the refinement of the proposed RoIs.
Classifiers are trained with an approximate bootstrapping
approach, called Minibootstrap~\cite{maiettini2019a}, which addresses the well-known issue of background-foreground class imbalance in object detection~\cite{oksuz2020}, while maintaining a short training time. In this work, the adoption of OOD permits to achieve a convenient speed/accuracy trade-off, since it allows to maintain a competitive accuracy with other DCNN-based approaches with a fraction of the optimization time required (seconds or minutes)~\cite{maiettini2019a,ceola2020a}.\\

\noindent{\bf{Scoring function.}} 
This function assigns a confidence score to the predictions for the images in the TARGET-UNLABELED. 
This score is then used by the AL and SS Selection policies to decide which images need to be manually annotated or can be considered as pseudo-ground truth. 
For this part, we employ the Cross-Image Validation (CIV) proposed in SSM~\cite{wang_towards_2018}. 
CIV stitches predicted image patches from the TARGET-UNLABELED on random images, sampled from TARGET-LABELED.
Then, it executes the detector on the stitched images and computes a \textit{consistency score} from the obtained confidence scores~\cite{wang_towards_2018}.\\

\noindent{\bf{AL and SS Selection policies}}.
Given the predicted detections obtained by the OOD and the \textit{consistency score} computed by the Scoring function, these two policies decide whether an image of the TARGET-UNLABELED is queried for annotation or the predicted detections are confident enough to be used for self-supervision.
Our main contribution relies on these last two components. 
Firstly, we target a stream-based scenario, since it is more suitable for on-line applications. 
Secondly, we consider a robotic setting with low annotation budget and a large domain shift of the TARGET-UNLABELED with respect to the TARGET-LABELED.
Specifically, for the \textit{AL Selection policy}, we consider several AL techniques, comparing their performance on the considered robotic setting and proposing a solution to enforce diversity during sampling.
The adopted AL baselines and the proposed solution are listed in Sec.~\ref{sec:methods:al}. 
Instead, for the \textit{SS Selection policy}, we consider a stream-based baseline and a novel strategy to overcome issues caused by the domain shift, both described in Sec.~\ref{sec:methods:ss}. 
Finally, another major difference with respect to previous work~\cite{maiettini2019b} is that we consider the case in which only one pass over the TARGET-UNLABELED data is allowed, while typically in standard Computer Vision, and also in~\cite{maiettini2019b}, an iterative process is used. This aspect is crucial for speeding up WSL. However, it makes detector refinement more challenging.

\subsection{Learning Protocol}
\label{sec:methods:learning}
The learning process is divided into: (i) \textit{Supervised phase} (represented by the light blue arrows in Fig.~\ref{fig:pipeline}), and (ii) \textit{Weakly-supervised phase} (represented by the orange arrows in Fig.~\ref{fig:pipeline}). Both phases rely on pre-trained Mask R-CNN's weights as feature extractor for the OOD. 
Those weights remain fixed, while model training and adaptation is performed by optimizing on the new data only the second stage of the OOD, i.e., (i)~the FALKON classifiers with the Minibootstrap technique and (ii) the RLS box-refinement model (see Sec.~\ref{sec:methods:overview} for details).
The \textit{Supervised phase} is performed within a few seconds of interaction with a human on the TARGET-LABELED, yielding a first detection model (the \textit{seed model}). In this phase the human shows the objects to the robot, handling them in their hand and annotations are automatically collected.
Then, in the \textit{WSL phase}, 
the SSL pseudo-ground truth and AL queries are selected from the TARGET-UNLABELED as described in Sec.~\ref{sec:methods:overview}, using the \textit{seed model}'s confidence scores.
Finally, they are added to the dataset which is used to re-train the on-line detector.

\subsection{Active Learning Strategies}
\label{sec:methods:al}
For AL selection, we considered both (i)~stream-based approaches, which are the focus of this work, being suited to robotic scenarios, and (ii)~pool-based ones.
\par
A simple, yet often effective, pool-based 
strategy is to sample uniformly at random the images with a confidence score below a threshold (\texttt{Uniform random} in Sec.~\ref{sec:experiments}).
Another diversity sampling strategy is to execute $k$-means clustering~\cite{hastie_elements_2009} on the image-level features and select the resulting cluster centers (\texttt{K-means-based AL} in Sec.~\ref{sec:experiments}).
In our analysis, we report results for both strategies.
\par
In stream-based AL settings, a simple selection strategy involves %unlabeled image 
confidence score thresholding followed by coin flipping~\cite{wang_towards_2018} for implementing uncertainty and diversity sampling, respectively (\texttt{coin-flip AL} in Sec.~\ref{sec:experiments}). 
Another, more practical,
solution is to exploit temporal coherence in image sequences to enforce sampling diversity~\cite{Aghdam_2019}.
Leveraging temporal coherence is particularly suitable for on-line robotic tasks, since data, coming in streams, needs to be acquired sequentially and is therefore highly temporally correlated.
To this aim, we consider the \texttt{Fixed temporal window} strategy, which employs a temporal window of fixed size $\Delta$ so that if frame $t$ is selected, any other frame within $[t - \Delta, t + \Delta] $ can no longer be considered for selection. 
While enforcing diversity, this strategy, by using a fixed $\Delta$, does not take into account: 
(i)~the exploration session duration, that is, the size $n_U$ of TARGET-UNLABELED, which might be known a-priori even in stream-based scenarios, and 
(ii)~the available manual annotation budget $k$.
 We show in Sec.~\ref{sec:exp-al} that this  results in poor performance for low $k$ when the TARGET-UNLABELED is redundant. 
\par
 To overcome this limitation, we propose to use an adaptive temporal window size, defined as 
$$\Delta_{n_U,k} = \frac{n_U \cdot \alpha}{k}$$
and referred to as \texttt{Adaptive temporal window} in Sec.~\ref{sec:experiments}.
This strategy allows to tailor the 
%tolerance 
strictness
(window size) of the temporal diversity-enforcing sampling to the overall amount of available unsupervised data $n_U$, while at the same time ensuring to make full use of the available budget $k$.
For instance, given a budget $k$, the adaptive window size grows linearly with $n_U$ in order to cover the entire duration of the exploration session.
$\alpha \in (0, k ) \subseteq \mathbb{R} $ is a hyperparameter accounting for the proportion of AL candidates with respect to $n_U$, which is unknown a priori.

%\par
%NOTE: here we could also expand on the denominator, which in this case is fixed to 2 but depends on: 1) the coin-flipping probability $p$ (do we report it somewhere as a hyperparameter?); 2) the proportion of AL candidates $r = n_{AL}/n_U$ (with $n_{AL}$ number of AL candidates).
%The full expression for the window size, assuming uniform distribution of the AL samples over the exploration session, would then be: 
%$$
%\Delta' = \frac{n_U \cdot r \cdot p}{k}
%$$

% \raf{To overcome this limitation}, we propose to use an adaptive temporal window size, \raf{defined as 
%$$\Delta_{n_U,k} = \frac{n_U}{2k}$$
%and referred to as \texttt{Adaptive temporal window} in Sec.~\ref{sec:experiments}.
%This strategy allows to tailor the 
%%tolerance 
%strictness
%(temporal window size) of the diversity-enforcing sampling to the overall amount of available unsupervised data $n_U$ while at the same time ensuring to make full use of the available budget $k$.
%For instance, given a budget $k$ the adaptive window size grows linearly with $n_U$ in order to cover the entire duration of the exploration session.
%\par
%NOTE: here we could also expand on the denominator, which in this case is fixed to 2 but depends on: 1) the coin-flipping probability $p$ (do we report it somewhere as a hyperparameter?); 2) the proportion of AL candidates $r = n_{AL}/n_U$ (with $n_{AL}$ number of AL candidates).
%The full expression for the window size, assuming uniform distribution of the AL samples over the exploration session, would then be: 
%$$
%\Delta' = \frac{n_U \cdot r \cdot p}{k}
%$$
%}
\begin{figure*}[t!]
	\centering
	\vspace{3mm}
	\includegraphics[width=0.8\textwidth]{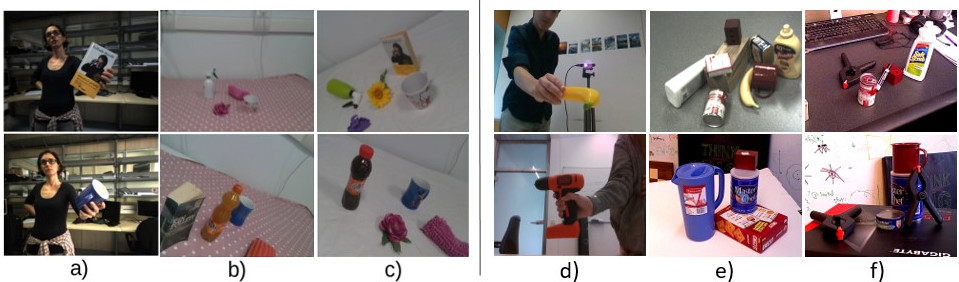}
	\caption{Example images of the datasets used for this work: \textbf{a)} \textsc{iCWT} dataset; \textbf{ b)} \textsc{pois} in TABLE-TOP dataset; \textbf{c)} \textsc{white} in TABLE-TOP dataset; \textbf{d)} HO-3D; \textbf{e)} YCB-Video training set, \textbf{f)} YCB-Video test set.}
	\label{fig:dataset}
\end{figure*}

\subsection{Semi-supervised Learning Strategies}
\label{sec:methods:ss}
For SS selection, we consider two stream-based baselines. 
The first is the \texttt{SS baseline}, which selects all the images passing CIV as pseudo ground truth.
 However, we show in Sec.~\ref{sec:exp-ssl} that under domain shift this leads to model degradation due to the abundance of false negatives.
 For this reason, in this work, we propose a more conservative strategy, namely \texttt{SS pos. only}, which only selects positive predictions and leaves out negative ones. 
 In Sec.~\ref{sec:exp-ssl}, we show that our approach 
 successfully counteracts severe model degradation. 
%  Moreover, in Sec.~\ref{sec:exp-pipeline} we show that integrating \texttt{SS pos. only} with AL further increases accuracy with the same labeling budget, improving the overall proposed system's efficiency.

\section{EXPERIMENTS}
\label{sec:experiments}
%%%%%%%%%%%%%%%%%%%%%%%%%%%%%%%%%%%%%%%%%%%%%%%%%%%%%%%%%%%%%%%%%%%%%%%%%%%%%%%%

% We now report on the experimental analysis to demonstrate the effectiveness of the proposed WSL techniques and compare them to the described baselines on the considered robotic scenario.

%\subsection{Datasets Description}
%\label{sec:exp:datasets}
%We consider two robotic datasets in our empirical evaluation. The first 
%is 
%iCubWorld Transformations~\cite{pasquale2019} 
%(iCWT),
%which contains images for 200 objects belonging to 20 categories (10 instances for each category).
%iCWT contains images of hand-held objects, demonstrated by a human teacher to the robot.
% Each object is acquired with different sequences representing specific viewpoint transformations: 
% 2D rotation (2D ROT), generic rotation (3D ROT), translation (TRANSL), scaling (SCALE) and 
% all transformations randomly combined (MIX) (see~\cite{pasquale2019}). 
%The second dataset is a collection of table-top sequences~\cite{maiettini_weakly_2019} 
%(TABLE-TOP),
%depicting 21 of the 200 objects from iCWT randomly placed on a table with two
%different tablecloths: 
%(i) pink/white pois (POIS) and 
%(ii) white (WHITE). 
%Refer to~\cite{maiettini_weakly_2019} for further details about the dataset. 
%Both datasets are publicly available\footnote{\url{https://robotology.github.io/iCubWorld/\#icubworld-transformations-modal/}}. 
%The two datasets contain the same objects, but with an important domain shift: 
%iCWT frames include the hand of the teacher,
%whereas TABLE-TOP
%has different backgrounds and depicts objects on a table.

The objective of our experiments is to evaluate the performance of the presented WSL techniques in improving detection performance under domain shift. Specifically, we consider the scenario of a robot previously trained with human interaction to detect handheld objects. We aim to generalize to a different setting (i.e., a table top) by exploiting the unlabeled data collected by the robot during autonomous exploration (\textit{Weakly-supervised phase} in Sec.~\ref{sec:methods:learning}).

\subsection{Experimental Setup}
\label{sec:exp:setup}
For the OOD, the weights of the Feature extractor are learned by training Mask R-CNN on the MS COCO~\cite{coco} dataset. ResNet50~\cite{He2015} has been considered as Mask R-CNN's convolutional backbone (we use the available pre-trained Mask R-CNN weights\footnote{\url{https://github.com/facebookresearch/maskrcnn-benchmark/blob/master/MODEL\_ZOO.md}}). During \textit{Supervised} and \textit{Weakly-supervised phases}, the feature extractor is fixed, while the FALKON classifiers and RLS are updated as explained in Sec.~\ref{sec:methods:learning} (we relied on~\cite{maiettini2019a} for hyper-parameters selection). This allows to achieve a training time of few seconds or minutes for each learning step.\\
Given the aforementioned target scenario, in our experiments we consider two different cases of domain adaptation from handheld (\textit{Supervised phase}) to table-top objects (\textit{Weakly-supervised phase}). Specifically, we adapt (i) from iCubWorld Transformations~\cite{pasquale2019} (iCWT) to a set of sequences depicting a subset of iCWT's objects on a table-top (TABLE-TOP) and (ii) from HO-3D~\cite{hampali2020} to YCB-Video~\cite{xiang2018}.\\

\noindent{\textbf{From iCWT to TABLE-TOP (iCubWorld domain)}}.
iCWT contains images for 200 handheld objects. Each object is demonstrated by a human teacher to the robot (as in~\cite{maiettini2017}) and is acquired with different sequences representing specific viewpoint transformations: 2D rotation (2D ROT), generic rotation (3D ROT), translation (TRANSL), scaling (SCALE) and all transformations (MIX) (see~\cite{pasquale2019}).
For the \textit{Supervised phase}, we employ a subset of the iCWT, considering 21 of the total 200 objects. All the transformations, except from MIX, are considered, resulting in a TARGET-LABELED of size $n_L \sim$6K. The TABLE-TOP depicts the same 21 objects randomly placed on a table with two different tablecloths: (i) pink/white pois (POIS) and (ii) white (WHITE). The two datasets contain the same objects, but with an important domain shift: iCWT frames include the hand of the teacher, whereas TABLE-TOP has different backgrounds and light conditions and depicts objects on a table. Refer to Fig.~\ref{fig:dataset} for a visual representation of the domain shift.
For the \textit{Weakly-supervised phase}, we consider the WHITE sequence as TARGET-UNLABELED while we leave the POIS sequence as test set to evaluate performance. These two sets are respectively of size $\sim$2K and $\sim$1K.
\\

\noindent{\textbf{From HO-3D to YCB-Video (YCB domain)}}. Similarly, in HO-3D and YCB-Video, objects from the YCB~\cite{calli2015} dataset are presented handheld by a human and in table-top sequences, respectively. Specifically, YCB-Video presents sequences for 21 objects while in HO-3D a subset of 9 of those objects are considered. Note that for our experiments we do not consider the labels for the remaining 12 objects in YCB-Video. For the \textit{Supervised phase}, we take from HO-3D at most four sequences for each object resulting in a TARGET-LABELED of size $n_L \sim$20K. For the \textit{Weakly-supervised phase}, we consider a set of $\sim$11.3K frames obtained by extracting one image every ten from the total 80 training video sequences available in the YCB-Video. As test set, instead, we consider the $\sim$3K keyframe~\cite{xiang2018} images chosen from the remaining 12 sequences in the YCB-Video.
\\

% When implementing AL queries on the TARGET-UNLABELED, we simulate human annotation for AL by fetching the actual ground truth from the dataset.
\noindent{\textbf{Evaluation metrics.}} We report performance in terms of mAP (mean Average Precision) at the IoU (Intersection over Union) threshold set to 0.5, as defined for Pascal VOC 2007 (see \cite{everingham_pascal_2010}). Specifically, we repeat each experiment for three trials and we present the results, reporting the mean and the standard deviation of the obtained accuracy\footnote{All experiments have been executed on a machine equipped with Intel Xeon E5-2690 v4 CPUs @2.60GHz, and an NVIDIA Tesla P100 GPU.}.

\begin{figure}[t]
	\centering
	\vspace{0.3cm}
	\includegraphics[width=0.3\textwidth]{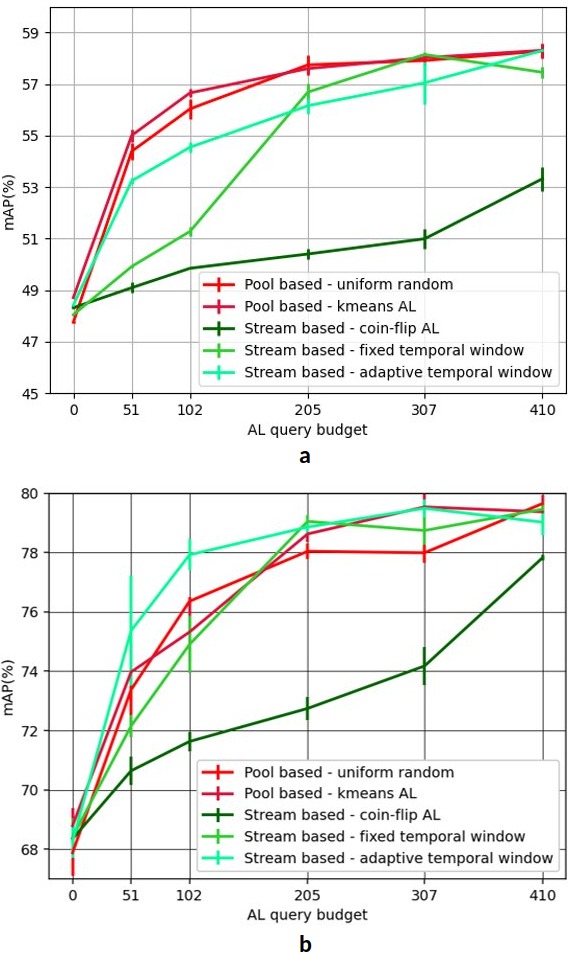}
	\caption{mAP comparison of pool-based (red) and stream-based (green) AL strategies with varying query budgets for iCubWorld (a) and YCB (b) domains.}
	\label{fig:al-baselines-comparison}
\end{figure}

\subsection{Active Learning Sampling Strategy Evaluation}\label{sec:exp-al}
In this section, we compare the AL techniques described in Sec.~\ref{sec:our-approach} considering different manual annotation budgets for both iCubWorld and YCB domains. 
To this aim, we report in Fig.~\ref{fig:al-baselines-comparison}a and b the mAP trends obtained by increasing the AL query budget during the \textit{Weakly-supervised phase}. 
Specifically, we report in red shades the performance obtained by the pool-based strategies (namely, \texttt{k-means-based AL} and \texttt{Uniform random} from Sec.~\ref{sec:our-approach}), and in green shades the stream-based ones (namely, \texttt{Coin-flip AL}, \texttt{Fixed temporal window}, and the proposed \texttt{Adaptive temporal window} from Sec.~\ref{sec:our-approach}).
We empirically set the fixed temporal window size as $\Delta=6$ and the adaptive temporal window hyperparameter as $\alpha = 0.5$ for the iCubWorld domain and  $\alpha = 0.4$ for the YCB domain.
As it can be observed in Fig.~\ref{fig:al-baselines-comparison}a, for iCubWorld domain the pool-based methods achieve the best mAP trends.
% We consider them as the upper bounds of  stream-based approaches. 
Notably, we observe that the \texttt{Uniform random} baseline is almost as effective as \texttt{$k$-means based-AL} and they both present an early steep slope for limited manual annotation budgets.
These two aspects are due to the fact that the considered TABLE-TOP dataset in the iCubWorld domain, contains sequences of similar (and thus redundant) frames which need to be properly filtered during data selection. 
This aspect of the dataset is also the main cause for the poor performance obtained by the two stream-based techniques: \texttt{Coin-flip AL} and \texttt{Fixed temporal window}, for low numbers of manual annotations. 
Indeed, while being more suited for a robotic application, by reasoning only on a frame-by-frame fashion, they lack global information on the whole data distribution, which turns out to be a critical drawback especially for limited manual annotation budgets. However, for higher budgets, the \texttt{Fixed temporal window} baseline achieves accuracies closer to the pool-based ones. Finally, the proposed \texttt{Adaptive temporal window} presents the best mAP trend, among the stream-based approaches, especially for low annotation budgets and it closely matches the pool-based ones.
On the contrary, as it can be observed in Fig.~\ref{fig:al-baselines-comparison}b, for the YCB domain the pool-based methods, the \texttt{Fixed} and the \texttt{Adaptive temporal window} present similar mAP trends. Specifically, the proposed \texttt{Adaptive temporal window} has the steepest slope. This is due to the fact that the YCB-Video dataset, differently from the TABLE-TOP, presents less redundant sequences and a smarter data selection based on temporal coherence provides the best performance.\\
Finally, it is important to note that the proposed \texttt{Adaptive temporal window} stream-based approach achieves significantly higher mAP values, than other stream-based techniques, for low annotation budgets for both domains. This makes it the most successful stream-based approach in such regime, which is the target of the presented work.
% \elisa{\begin{table}[t]
% 	\centering
% 	\vspace{0.2cm}
% 	\caption{\texttt{SS baseline} results for large ($1^{st}$ row) and small ($2^{nd}$ row) domain shift between supervised and unsupervised datasets.}
% 	\begin{tabular}{l|l|l|l|}
% 		\cline{2-4}
% 		& \multicolumn{1}{c|}{\textbf{\begin{tabular}[c]{@{}c@{}}Supervised phase\\ (mAP(\%))\end{tabular}}} & \multicolumn{1}{c|}{\textbf{\begin{tabular}[c]{@{}c@{}}SS baseline\\ (mAP(\%))\end{tabular}}} & \multicolumn{1}{c|}{\textbf{\begin{tabular}[c]{@{}c@{}}SS samples\\ percentage\end{tabular}}} \\ \hline
% 		\multicolumn{1}{|l|}{\textbf{Large DS}} & 48.8 $\pm$ 0.3       &        37.9 $\pm$ 1.8             &      $\sim$ 12\%          \\ \hline
% 		\multicolumn{1}{|l|}{\textbf{Small DS}} &         40.3 $\pm$ 0.9             &        47.3 $\pm$ 0.9              &      $\sim$     35\%         \\ \hline
% 	\end{tabular}
% 	\label{table:SS_baseline}
% \end{table}}

\begin{table}[t]
	\vspace{0.3cm}
	\centering
	\caption{Results obtained by \texttt{SS baseline} ($2^{nd}$ column) and \texttt{SS positives} ($3^{rd}$ column) for large ($1^{st}$ row) and small ($2^{nd}$ row) domain shift from the supervised ($1^{st}$ column) to the weakly-supervised phase.}
	\resizebox{8.7cm}{!}{
	    \begin{tabular}{c|c|c|c|c|}
			\cline{2-5}
			& \multicolumn{1}{c|}{\textbf{\shortstack{Sup. phase \\ (mAP(\%))}}} & \multicolumn{1}{c|}{\textbf{\shortstack{SS baseline \\ (mAP(\%))}}} & \multicolumn{1}{c|}{\textbf{\shortstack{SS pos. only \\ (mAP(\%))}}} & \multicolumn{1}{c|}{\textbf{\shortstack{SS \\ samples}}} \\ \hline
			\multicolumn{1}{|l|}{\textbf{Large DS}} & 48.8 $\pm$ 0.3       &        37.9 $\pm$ 1.8             &         50.9 $\pm$ 0.06             &      $\sim$ 12\%          \\ \hline
			\multicolumn{1}{|l|}{\textbf{Small DS}} & 40.3 $\pm$ 0.9       &        47.1 $\pm$ 0.1             &         46.6 $\pm$ 0.2             &      $\sim$ 35\%         \\ \hline
		\end{tabular}
	}
	\label{table:SS_baseline}
	\vspace{-4mm}
\end{table}

\subsection{Semi-supervised Learning Evaluation}\label{sec:exp-ssl}

In this section, we investigate the impact of domain shift from handheld objects to table-top datasets when no labeling is allowed (i.e., SSL). 
To this aim, we report in Tab.~\ref{table:SS_baseline} the results of applying the \texttt{SS baseline} (as defined in Sec.~\ref{sec:our-approach}) in the two following scenarios:
\begin{itemize}
	\item \textbf{Large domain shift}. In this case (\textit{Large DS} row in Tab.~\ref{table:SS_baseline}), we consider the scenario in which the TARGET-UNLABELED presents a completely different setting (i.e., a table top) with respect to the TARGET-LABELED (i.e., hand-held). 
To this end we used the two datasets described in Sec.~\ref{sec:exp:setup}.
	\item \textbf{Small domain shift}. In this case (\textit{Small DS} row in Tab.~\ref{table:SS_baseline}), TARGET-LABELED and TARGET-UNLABELED present similar conditions. 
The only difference in the latter one is that the objects are presented, unlabeled, with different view poses. 
To this aim, we considered as TARGET a 30-object identification task from iCWT. 
For each object, we then use the TRANSL sequence ($\sim$2K
	images) as TARGET-LABELED and the union of the 2D ROT, 3D ROT, and SCALE sequences ($\sim$6K images) as the TARGET-UNLABELED. We test on the MIX sequences of all the objects ($\sim$4.5K images).
\end{itemize}
Note that, in this experiment, we use the iCubWorld domain only because the explicit sub-division in different viewpoint transformations of iCWT allows to control the dataset split in TARGET-LABELED and TARGET-UNLABELED such that they present similar, but not identical, conditions. This allows to precisely identify the \textit{Small DS} setting.

\begin{figure}[t]
	\centering
	\vspace{0.3cm}
	\includegraphics[width=0.35\textwidth]{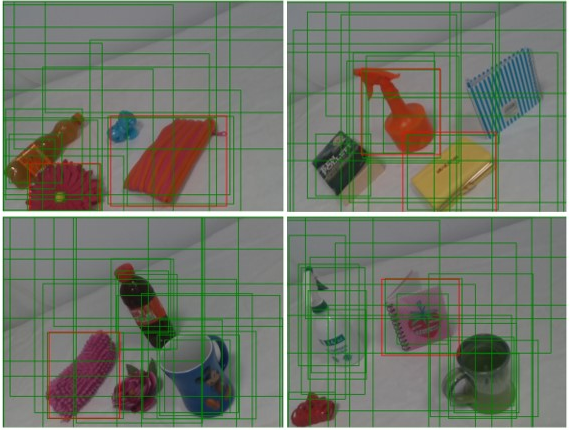}
	\caption{Example predictions on the TARGET-UNLABELED before model adaptation, selected by the \texttt{SS Baseline} for SS as positives (red boxes) and negatives (green boxes).}
	\label{fig:ss_failure}
\end{figure}

Tab.~\ref{table:SS_baseline} reports the results obtained in both cases. 
For each row, we report the mAP (represented as mean and standard deviation of the different repetitions) after the \textit{Supervised phase} (first column) and after the \textit{Weakly-supervised phase} for both \texttt{SS baseline} and \texttt{SS pos. only} (second and third columns). 
Moreover, in the fourth column we report the average percentage of samples selected by the SS process over the total. 
As it can be observed, adding self-supervised data with \texttt{SS baseline}, with small domain shift, results in an improvement in accuracy.
On the contrary, with a larger domain shift, it leads to a significant accuracy deterioration. 
A reason for this phenomenon can be identified by analyzing the pseudo-ground truth generated by the SS process. 
We report in Fig.~\ref{fig:ss_failure} some representative images depicting in green the region proposal candidates classified as background by the detection system and that are therefore added as negative samples to the dataset by the \texttt{SS baseline}.
The actual detections which instead are considered as positive samples in the SS process are shown in red. 
It can be noticed that, with large domain shift, only few objects are correctly detected and therefore added to the training set as positives, while most others are false negatives which are automatically annotated as background samples.
Clearly, retraining the detection model with such a poorly-labeled dataset leads to the sharp performance decay shown in Tab.~\ref{table:SS_baseline}. This confirms similar findings from the literature~\cite{li2021}, in the considered setting. Note that, we empirically noticed that lowering the confidence threshold used to determine a positive prediction is not suitable since, while not ensuring less false negatives, it leads to imprecise predictions, with a similar negative effect on the subsequent training.
%
% We propose a solution to this problem by introducing the \texttt{SS pos. only} strategy described in Sec.~\ref{sec:our-approach}.
% According to this more conservative strategy, only the detections predicted as positive
% are included in the self-supervised dataset. 
% The other regions  are filtered away, since they might contain numerous false negatives (i.e., objects of interest wrongly classified as background by the current model). 
% We report the results obtained with the proposed approach in Tab.~\ref{table:SS_only_pos}.
% The proposed strategy, not only effectively removes wrong labels from the dataset, recovering from the two-digits accuracy decay, but also  successfully yields a significant improvement in performance. Moreover, it allows to drastically decrease the standard deviation of the obtained accuracy, from 1.8\% to 0.06\%, demonstrating to be less sensitive to statistical fluctuations. Finally, in the next section, we show that it can be used to further improve data efficiency of AL techniques.
In Sec.~\ref{sec:our-approach}, we introduce the \texttt{SS pos. only} to address this issue. This more conservative strategy includes only the regions predicted as positive in the SS dataset, while the others are filtered away, avoiding adding false negatives. Third column in Tab.~\ref{table:SS_baseline} shows that this strategy, does not modify the basline in case of Small DS where SS data is already reliable. However, for Large DS, not only effectively removes wrong labels from the dataset, recovering from the two-digits accuracy decay, but also successfully yields a performance improvement of $\sim$2 points. Moreover, it allows to drastically reduce the standard deviation of the obtained accuracy, from 1.8\% to 0.06\%, being less sensitive to statistical fluctuations. This demonstrates that, in cases when no human manual annotation is allowed, a robot trained to detect handheld objects can explore the new domain, self-annotating the newly collected data and improving detection performance.

\section{CONCLUSIONS}
\label{sec:conclusions}
%%%%%%%%%%%%%%%%%%%%%%%%%%%%%%%%%%%%%%%%%%%%%%%%%%%%%%%%%%%%%%%%%%%%%%%%%%%%%%%%

In this paper, we target the scenario of a robot trained with human interaction to detect handheld objects, aiming to improve detection performance in different settings with autonomous exploration and limited human intervention. We empirically demonstrate that general purpose WSL techniques are unsuitable for challenging robotic scenarios and we propose solutions to both (i) enforce diversity sampling for AL queries and (ii) improve strong positives selection for SSL under severe domain shift. Finally, we build on previous work~\cite{maiettini2019b}, presenting and empirically evaluating a stream-based weakly-supervised on-line object detection pipeline for Robotics, which exploit the robot interaction with the environment and the human teacher to update and improve performance of the visual system. It significantly alleviates the annotation burden for on-line model adaptation to novel settings while maximizing accuracy.
% present and empirically evaluate a stream-based weakly-supervised on-line object detection pipeline for robotics, improving on previous work~\cite{maiettini2019b}. 
%We compare several AL and SS techniques, proposing solutions to overcome stream-based AL limitations and performance degradation in SSL due to domain shift.
%
%Reliable perception and efficient adaptation to novel tasks and conditions are prominent skills for robots to operate in ever-changing environments. 
%Fast optimization times and annotated training data efficiency are critical aspects for learning-based object detection systems in robotics. 
%From this perspective, this work proposes novel solutions to significantly alleviate the annotation burden for on-line model adaptation to novel settings while maximizing accuracy, for enabling robots to explore the environment more autonomously.

%%%%%%%%%%%%%%%%%%%%%%%%%%%%%%%%%%%%%%%%%%%%%%%%%%%%%%%%%%%%%%%%%%%%%%%%%%%%%%%%

\bibliographystyle{IEEEtran}
\bibliography{IEEEabrv,eal,bibliography}

\end{document}